# A FREE LUNCH FROM VIT: ADAPTIVE ATTENTION MULTI-SCALE FUSION TRANSFORMER FOR FINE-GRAINED VISUAL RECOGNITION

*Yuan Zhang[1], Jian Cao[1], Ling Zhang[1], Xiangcheng Liu[1], Zhiyi Wang[1], Feng Ling[1], Weiqian Chen[1,2]*

[1] School of Software and Microelectronics, Peking University, China.
[2] Alibaba Group, Inc.

## ABSTRACT

Learning subtle representation about object parts plays a vital role in fine-grained visual recognition (FGVR) field. The vision transformer (ViT) achieves promising results on computer vision due to its attention mechanism. Nonetheless, with the fixed size of patches in ViT, the class token in deep layer focuses on the global receptive field and cannot generate multi-granularity features for FGVR. To capture region attention without box annotations and compensate for ViT shortcomings in FGVR, we propose a novel method named Adaptive attention multi-scale Fusion Transformer (AFTrans). The Selective Attention Collection Module (SACM) in our approach leverages attention weights in ViT and filters them adaptively to correspond with the relative importance of input patches. The multiple scales (global and local) pipeline is supervised by our weights sharing encoder and can be easily trained end-to-end. Comprehensive experiments demonstrate that AFTrans can achieve SOTA performance on three published fine-grained benchmarks: CUB-200-2011, Stanford Dogs and iNat2017.

## 1. INTRODUCTION

Compared with generic object recognition, fine-grained visual recognition is a challenging task, with large intra-class variance and small inter-class variance. Fine-grained objects (birds [1] and dogs [2]) are similar at a cursory glance, while they can be recognized by details in discriminative local parts.

Methods for FGVR can be classified into three categories: feature-encoding methods, localization-based methods and attention-based methods. Compared with feature-encoding methods [3, 4], the localization methods can explicitly tell the subtle differences among different sub-classes and usually yields better results. Recent localization works [5, 6, 7] integrate region proposal networks (RPN) to propose bounding boxes containing details, while the early [8, 9, 10] rely on the annotations of parts where extra labor is necessary. However, RPN-based methods ignore the relationships among regions they selected. Attention-based [11, 12, 13] methods select region attentions in images by exploiting the attention properties of the feature maps of CNN itself , releasing the reliance

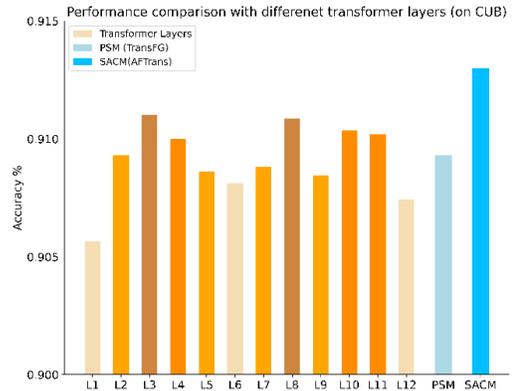

**Fig. 1**. The performance comparison on CUB-200-2011 with single or multiple fusion attention weights in transformer layers, utilized for proposing the critical parts.

on manually annotation. When it comes to attention mechanisms, transformer comes to our mind.

Transformer in CV field has become a research hotspot, with its application on image classification [14] and image detection [15]. The innovation of image sequentialization in vision transformer, where ViT flattens the image patches and transforms them into patch tokens, inspires the community to utilize the innate attention mechanism. Nonetheless, native vision transformer cannot play to its strengths on FGVR directly. For example, the receptive field of ViT cannot be effectively extended, since the length of patch tokens does not change as its encoder blocks increases. Besides, the model may not effectively capture the region attention carried in the patch tokens. To tackle above problems, TransFG [16] proposed to exclude immaterial inputs of the final transformer layer with ViT inherent attention weights to choose which tokens to stay. However, when generating the attention map for picking up tokens, TransFG cannot use all tranformer layers attention completely. As shown in the Fig1, better accuracy can be achieved with attention in single transform layer (e.g. L3) compared with PSM in TransFG, while SACM achieves the best with adaptive integration. We prove that attention weights in each layer do not play the equal role in terminating the fused attention map, which aims to correspond to the relative importance of input tokens. Besides, TransFG over-looks the supervise for global loss and lacks the combination of multi-scale and multi-branch.

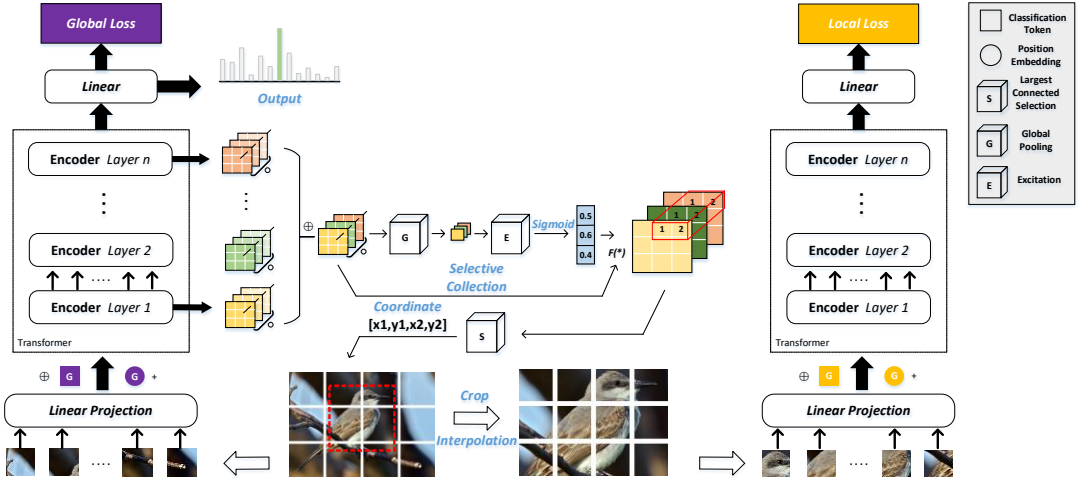

**Fig. 2.** The Architecture of AFTrans. Images are split into patches and sent into ViT. Collected from attention weights processed by Hadamard product in each transform layer, the Selective Attention Collection Module (SACM) combines with critical tokens and converts it into the coordinates of local regions like the red bounding box. Notably, the linear projection, the transformer layers, and linear layer are parameter-sharing, while classification tokens and position embeddings do not. Best viewed in color.

To this end, we launch the adaptive attention multi-scale fusion transformer, which leverages the transformer innate attention mechanism to strengthen locality in a multi-scale pipeline, which is shown in the Fig2. All told, how to digest and absorb 'the free lunch' from ViT better is our work focus.

## 2. PROPOSED METHOD

### 2.1. Raw Materials: Attention in Vision Transformer

Vision transformer successfully applies transformer to the visual field by introducing patch embedding. For the sequence patch embedding input of changing the image into the corresponding 1D, 2D image $x \in \mathbb{R}^{H \times W \times C}$, is divided into $N$ patches with size $P \times P$, $x_p \in \mathbb{R}^{N \times (P \times P \times C)}$. Patches are passed through convolution layer with kernel size $P \times P$, and then added with the randomly initialized learnable position embedding to obtain the 1D sequence. The formula is as follows:

$$x_{embedded} = [PaE(x_{p1}, x_{p1}, \ldots, x_{pN})] + PosE \quad (1)$$

Where $x_{embedded}$ is the embedding value before entering the encoder, $PaE$ is patch embedding, and $PosE$ is position embedding. Position embedding can bring corresponding location information to patches without location information. It should be noted that $x_{p1}$ is a randomly initialized learnable class token. Transformer encoder is composed of multi-head self-attention (MSA) and multi-layer perceptron (MLP) blocks. The output of $l$-th layer is shown below:

$$z'_l = Softmax(Q * K^T) * V \quad (2)$$

$$z_l = z'_l * W_{MLP} + z'_l \quad (3)$$

Where $Q, K, V$ are Query, Key and Value vectors respectively. $z'_l$ is the output passing through MSA but not MLP yet, and $z_l$ is the output of the $l$-th layer. Vision transformer selects the first output generated from the class token for classification tasks. This is because class token has more global rich information than other patches. The formula of classification network is as follows:

$$out = Softmax(z_{cls} \cdot W_{cls}) \quad (4)$$

Where $out$ is the output of the classification network, $W_{cls}$ is the weight of net, and $z_{cls}$ is the output of the encoder corresponding to class token.

### 2.2. Cooking Method: AFTrans Model

The Vision Transformer can be for fine-grained classification, while it does not well capture the local information required for FGVR. We propose the adaptive attention multi-scale fusion transformer (AFTrans) to cope with the above problem.

*2.2.1 Selective Attention Collection Module*

To capture the local discriminative part $e_2$ of the raw image $e_1$, the SACM collects multi-head self-attention weights $a_l$ of each transformer layer, and adaptively finetunes their perception contribution to attention map. We first take out the attention weights of each transformer layer as:

$$A = softmax\left(\frac{QK^T}{D^{\frac{1}{2}}}\right) = [a_0, a_1, \ldots, a_{L-1}] \quad (5)$$

$D$ is the dimension of embedding space. We argue that effective attention features can be gradually accumulated and amplified in each layer by the Hadamard product, so we integrate each layer's attention weights of $K$ multi-heads as:

$$A_i = \bigodot_{j=0}^{K-1} a_i^j \quad i \in (0,1,\ldots,L-1) \quad (6)$$

$$a_i^j = [a_i^0, a_i^1, \ldots, a_i^{C-1}] \quad j \in (0,1,\ldots,K-1) \quad (7)$$

Then we concatenate attention weights of $L$ layers as $W_c$. The core of SACM is to generate 'attention in attention'. We first adopt global average-pooling to squeeze the token dimension of the input attention weights. Descriptors are forwarded to a network to produce our layer attention $M_L \in \mathbb{R}^{N \times (1 \times 1 \times L)}$. The network is composed of multi-layer perceptron (*MLP*) with one hidden layer, whose size is set to $\mathbb{R}^{C/r \times 1 \times 1}$, where $r$ is the reduction ratio. Then we employ a gating mechanism with sigmoid activation $\sigma$. The fused attention map $A_F$ is computed as:

$$A_F = M_L * W_c \qquad (8)$$

$$M_L = \sigma(MLP(Pool(\bigoplus_{i=0}^{L-1} a_i))) \qquad (9)$$

$$= \sigma(W_2 \delta(W_1 * W_c^{pool}))$$

Where $\delta$ is ReLU, $W_1 \in \mathbb{R}^{C/r \times C}$ and $W_2 \in \mathbb{R}^{C \times C/r}$. The matrix $A_F$ reflects the correlation between image tokens. In order to select effective tokens for classification, we choose the most relevant $L \times N \times \lambda$ tokens with class token for classification as $\tilde{A}_F$, and map them into coordinates in $e_1$. Where $N$ is the number of patches and $\lambda$ is the threshold of cropping. Then we employ *the max connected region search* algorithm to extract the largest connected component of $\tilde{A}_F$ for localizing and zooming region attention in the raw image $e_1$.

*2.2.2 Multi-scale Fusion Training Pipeline*

To improve classification ability and robustness for images of different scales, we take multi-granularity as our general framework, which is inspired by MMAL and RAMS-Trans [30]. First, send the raw input image $e_1$ into ViT and get global loss. Then, get the local image $e_2$ cropped from $e_1$ and resized $224 \times 224$ by bilinear interpolation, and feed it into the enocder with its own cls tokens to get local loss. The two branches share one ViT to avoid more computation. During training, our loss function is a multi-task loss with global and local loss, supervised by cross entropy loss respectively:

$$Loss = \alpha Loss_g + \beta Loss_l \qquad (10)$$

$\alpha$ and $\beta$ are the coefficients complementary to each other. $Loss_g$ represents the classification loss of general scale and $Loss_l$ is the guided loss designed to guide the SACM to select the more discriminative parts. It enables the convergent model to make predictions based on overall structural characteristics of the object and the characteristics of the region part. When the model goes through the inference phase, we prefer taking global logits to decide the classification result.

## 3. EXPERIMENTS CONFIGURATIONS

### 3.1. Datasets

We evaluate our proposed AFTrans on widely recognized fine-grained benchmarks, namely CUB-200-2011, Stanford Cars, Stanford Dogs and iNat2017 [29]. We only utilize the image labels provided by these datasets with no extra annotations.

### 3.2. Implementation Details

In all experiments, we first resize raw images to size 448×448 for global branch and part images to size 224×224 for local branch. We load weights from the official ViT-B_16 model pre-trained on ImageNet21k. We split images to patches of size 16×16 and the step size of sliding window is set to be 12. We employ the SGD optimizer to optimize with a mini-batch size of 12. We use weight decay 0. We take cosine annealing to adjust the learning rate and the first 500 steps are warm-up. The coefficients in Eq13 are set to be 1 and 1. The hyperparameter $\lambda$ is chosen to be 0.4. All the experiments are performed on four Quadro RTX 8000 GPUs with Pytorch as our code-base. APEX with FP16 training is necessary.

## 4. RESULTS AND VISUALIZATION

### 4.1. Performance Comparison

We compare AFTrans with SOTA works on above mentioned fine-grained benchmarks. The experiment results on CUB-200-2011, Stanford Cars and Stanford Dogs are shown in tab1. From the results, our method AFTrans outperforms the strong baseline with a margin on CUB dataset, Stanford Dogs and achieves competitive performance on Stanford Cars in Transformer series.

**Table 1.** Comparison of different SOTA methods on CUB-200-2011, Stanford Cars and Stanford Dogs (%).

| Method | Backbone | CUB | CAR | DOG |
|---|---|---|---|---|
| RA-CNN [18] | VGG-19 | 85.3 | 92.5 | 87.3 |
| MaxEnt [19] | DenseNet-161 | 86.6 | 93.0 | 83.6 |
| DFL-CNN [20] | ResNet-50 | 87.4 | 93.1 | 84.9 |
| Cross-X [21] | ResNet-50 | 87.7 | 94.6 | 88.9 |
| SEF [28] | ResNet-50 | 87.3 | 94.0 | 88.8 |
| FDL [24] | DenseNet-161 | 89.1 | 94.2 | 84.9 |
| PMG [22] | ResNet-50 | 89.6 | 95.1 | - |
| MMAL [16] | ResNet-50 | 89.6 | 95.0 | - |
| API-Net [23] | DenseNet-161 | 90.0 | **95.3** | 90.3 |
| StackedLSTM [7] | GoogleNet | 90.4 | - | - |
| ViT [14] | ViT-B_16 | 90.2 | 93.5 | 91.2 |
| TransFG [16] | ViT-B_16 | 90.9 | 94.1 | 90.4 |
| AFTrans (ours) | ViT-B_16 | **91.5** | 95.0 | **91.6** |

**Table 2.** Comparison of SOTA methods on iNaturalist2017.

| Method | Backbone | Acc.(%) |
|---|---|---|
| ResNet-152 [17] | ResNet-152 | 59.0 |
| SSN [25] | ResNet-101 | 65.2 |
| IARG [26] | ResNet-101 | 66.8 |
| IncResNetV2 [27] | IncResNetV2 | 67.3 |
| TASN [5] | ResNet-101 | 68.2 |
| ViT [14] | ViT-B_16 | 68.0 |
| TransFG & PSM [16] | ViT-B_16 | 67.4 |
| AFTrans (ours) | ViT-B_16 | **68.9** |

For CUB-200-2011, almost all the FGVR methods evaluate their model on it. Compared to the best result TransFG so far, AFTrans achieves 0.6% improvement and outperforms all CNN-based methods. While base framework ViT achieves good performance on CUB, with the addition of our SACM and multi-branch, it achieves a further **1.3%** improvement.

Our method achieves well-matched result in Transformer series and gets **1.5%** improvement compared to baseline ViT

on Stanford Cars. Due to less necessary work on locatization with simpler backgrounds, AFTrans is lightly worse than PMG and API-Net. For Stanford Dogs, our approach is better than TransFG with its PSM by 1.2%. AFTrans gets **0.4%** improvement compared to ViT, attributed to its adaptive attention mechanisms catching subtle differences between species.

Tab2 shows our evaluation results on iNat2017, which is a large-scale FGVR dataset. Few methods tested on iNat2017 because of the complicate background and computational complexity. Notably, baseline ViT is born for large scale datasets, and outperformes ResNet152 by 9.0%. Based on ViT, our method AFTrans is can achieve an improvement of **0.9%** and outperforms all the SOTA methods.

### 4.2. Ablation Studies

We conduct ablation studies to understand variants in AFTrans architecture, which are done on CUB-200-2011 dataset.

**Impact of Selective Attention Collection Module.** Shown in tab3, the SACM making a better performance than ViT by proposing subtle parts, which improves by **1.1%**. Combined with Fig1, we argue that SACM adaptively fuses perception of attention contribution and forces the attention map $A_F$ to learn the relative importance of input patches.

**Table 3.** Ablation experiment on attention weights fusion method.

| Methods | Acc.(%) |
|---|---|
| None(ViT) | 90.2 |
| PSM(TransFG) | 90.9 |
| SACM(AFTrans) | **91.3** |

**Impact of branch logits in multi-scale framework.** There are global branch and local branch in AFTrans, so whose logits decide the classification result needs confirmation. Shown in the tab4, decided by the global logits gets the best result, which observes the object in a holistic perspective and is the resource of attention weights.

**Table 4.** Ablation experiment on branch logits ($\lambda = 0.4$).

| Global Logits | Local Logits | Acc.(%) |
|---|---|---|
|  | ✓ | 90.6 |
| ✓ |  | **91.5** |
| ✓ | ✓ | 91.0 |

**Impact of hyperparameters $\lambda$.** $\lambda$ controls SACM to accept which tokens are effective contribution to generate the coordinate. The best accuracy occurs when $\lambda$ is set to 0.4. We conclude that when $\lambda$ is small, the SACM will crop fewer in the raw images, losing many critical regions to some extent. While when $\lambda$ is large, the SACM will crop more in the raw images, resulting in redundant regions fed into model again.

**Table 5.** Ablation experiment on thresh $\lambda$.

| Value of $\lambda$ | 0.1 | 0.2 | 0.3 | 0.4 | 0.5 |
|---|---|---|---|---|---|
| Acc.(%) | 90.6 | 90.9 | 91.3 | **91.5** | 91.1 |

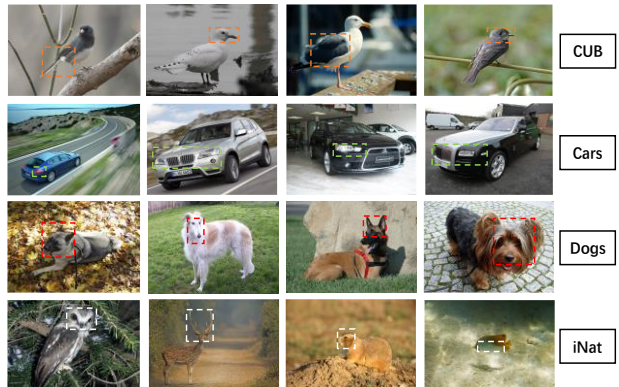

**Fig.3.** Visualization of discriminative local parts learned by the SACM on four FGVR datasets. Best viewed in color.

**Impact of the manner of generating local coordinate.** How to convert $L \times N \times \lambda$ coordinates to the last coordinate needs discussion. One is choosing extreme values of coordinates converted as the boundary points, the other is extracting the largest connected component (LCC) of coordinates. Tab6 confirms that LCC method achieves better performance, while it ignores immaterial coordinates forwardly.

**Table 6.** Ablation experiment on generating the coordinate of local regions

| Methods | Acc.(%) |
|---|---|
| Extreme Values | 90.9 |
| LCC | **91.5** |

### 4.3. Visualization Analysis

Fig3 conveys that local regions the AFTrans proposed do contain richer fine grained information. The bounding boxes are obtained from attention map fused by SACM. We can clearly see that AFTrans has captured the most discriminative regions of birds, such as head, wings and tail. For Cars, headlights and radiator grille may be the key points, while for dogs, head and ears are critical. In complex dataset iNat2017, AFTrans still can select local parts for an object, i,e., heads for Aves, horns and head for Mammalia; fins for Actinopterygii.

### 5. CONCLUSIONS

In this paper, we propose a novel method named adaptive attention multi-scale fusion transformer (AFTrans) to learn subtle region attention without box annotations and compensate for ViT shortcomings in FGVR. The multi-scale structure can make full use of the images obtained by SACM to achieve excellent performance. Our algorithm can achieve SOTA performance on three fine-grained benchmarks: CUB-200-2011, Stanford Dogs and iNat2017. The future work is how to locate region attention more precisely to further improve the classification accuracy.